\documentclass{article}
\usepackage{PRIMEarxiv}
\pdfoutput=1
\usepackage[utf8]{inputenc} 
\usepackage[T1]{fontenc}    
\usepackage{hyperref}       
\usepackage{url}            
\usepackage{booktabs}       
\usepackage{amsfonts}       
\usepackage{nicefrac}       
\usepackage{microtype}      
\usepackage{lipsum}
\usepackage{fancyhdr}       
\usepackage{graphicx}       
\usepackage[nocompress]{cite}

\graphicspath{{media/}}     

\usepackage[american]{babel}

\usepackage{mathtools} 
\usepackage{tikz} 
\usepackage{times}
\usepackage{soul}
\usepackage{amsmath}
\usepackage{array}
\usepackage{amsthm}
\usepackage{algorithm}
\usepackage{algorithmic}
\usepackage[switch]{lineno}
\usepackage{subcaption}
\usepackage{multirow}
\usepackage{svg}
\usepackage{amssymb}  
\usepackage{amsfonts}
\usepackage{caption}
\usepackage{float}
\usepackage{indentfirst}
\usepackage{mathtools}
\usepackage{algorithm, algorithmic}

\title{STCL:Curriculum learning Strategies for deep learning image steganography models
}
\author {
	FengChun Liu$^{1}$, Tong Zhang$^{2}$, Chunying Zhang$^{3}$\\
	$^{1}$Qianan College, North China University of Science and Technology, Tangshan, Hebei 063210, China\\
	$^{2}$School of Cyberspace Security, Beijing University of Posts and Telecommunications, Beijing 100876, China\\
	$^{3}$College of Science, North China University of Science and Technology, Tangshan, Hebei 063210, China\\
	\texttt{lnobliu@ncst.edu.cn, zenozt@bupt.edu.cn, hblg\_zcy@126.com}
}

\begin{document}
\maketitle

\begin{abstract}
Aiming at the problems of poor quality of steganographic images and slow network convergence of image steganography models based on deep learning, this paper proposes a Steganography Curriculum Learning training strategy (STCL)  for deep learning image steganography models. So that only easy images are selected for training when the model has poor fitting ability at the initial stage, and gradually expand to more difficult images, the strategy includes a difficulty evaluation strategy based on the teacher model and an knee point-based training scheduling strategy. Firstly, multiple teacher models are trained, and the consistency of the quality of steganographic images under multiple teacher models is used as the difficulty score to construct the training subsets from easy to difficult. Secondly, a training control strategy based on knee points is proposed to reduce the possibility of overfitting on small training sets and accelerate the training process. Experimental results on three large public datasets, ALASKA2, VOC2012 and ImageNet, show that the proposed image steganography scheme is able to improve the model performance under multiple algorithmic frameworks, which not only has a high PSNR, SSIM score, and decoding accuracy, but also the steganographic images generated by the model under the training of the STCL strategy have a low steganography analysis scores.

You can find our code at \href{https://github.com/chaos-boops/STCL}{https://github.com/chaos-boops/STCL}.

\end{abstract}



\section{Introduction}\label{sec:intro}
With the rapid development of network technology, network information security has become a crucial issue today. Especially, multimedia data, because of its richer expression and wider use, has more information leakage and privacy damage in the process of acquisition, modification and dissemination, how to guarantee the security of multimedia information has gradually become an important research topic in the field of cyberspace security. Steganography is a technology that protects the secure transmission of secret information by hiding information in different media that cannot be recognized by human vision, and is applied to private communication, military, industry and other commonly used scenarios that require the protection of confidential data. As an important method to realize covert communication, steganography has become a popular research direction in the field of information security.

Steganography embeds secret information into a carrier (image, text, audio, video, etc.) through a specific encoding algorithm, and then the receiver of the information realizes the extraction of the secret information through a specific decoding algorithm. Steganography is divided into traditional steganography algorithms \cite{hameed2023secure,bavrina2022investigation} and deep learning based steganography algorithms \cite{zhang2019steganogan,zhang2019invisible}. Traditional steganography algorithms are simple in principle and have low embedding costs, but are prone to visual artifacts and problems such as “value pair” effects and excessive changes in statistical features, such as least significant bit substitution \cite{hameed2023secure}. In recent years, with the introduction of deep learning into the field of steganalysis, the detection accuracy of steganography has been rapidly improved and the training time of the model has been reduced, while the traditional image steganography scheme is unable to resist the detection of steganography based on deep learning.

Generative Adversarial Network (GAN) \cite{goodfellow2020generative} proposed by Goodfellow in 2014 provides an opportunity to combine image steganography with deep learning networks. steGAN \cite{hayes2017generating} steganographic model proposed by Hayes et al. defines a three-way adversarial game of encoding, decoding and steganalysis, which opens up a new research direction in the field of steganography, whose steganographic images can deceive steganalysis networks.

Subsequent researchers have focused on the three indicators of steganographic quality, capacity, and security of steganographic images, concentrating on the use of adversarial training for steganographic tasks to enhance the security of steganographic images \cite{wang2018sstegan}, and improving the coding network to enhance the quality of steganography and other work. However, existing steganography research is prone to image distortion and artifacts as the steganographic capacity expands, and there are also problems such as reduced information extraction accuracy. Existing deep learning-based steganography schemes treat the samples uniformly during the model training process and use a randomly disrupted training strategy, resulting in poor performance of the steganography model on test images such as those containing some solid color regions.

Curriculum learning \cite{bengio2009curriculum}, which mimics the basic idea of step-by-step progression of humans in the process of learning a curriculum, is a model training strategy for non-convex optimization that advocates that the model learns in the order of easy samples to more difficult samples, and has been widely used in areas such as computer vision, natural language processing, and reinforcement learning. Curriculum learning is regarded as a continuation method for global optimization of non-convex functions, and it is believed that Curriculum learning is effective because it is able to spend less time on noisy and training-difficult data at the beginning of the training period, while at the same time it can guide the training towards a better local optimum and a better generalization effect.

Inspired by curriculum learning, a training strategy for deep learning image steganography models is proposed, including a difficulty assessment strategy based on teacher models and an inflection point-based training scheduling strategy. Specifically, three teacher models with different levels are trained individually, and the consistency of the quality of steganography under the three teacher models for each sample is used as the difficulty score to construct the training subsets from easy to hard. At the beginning of training, only the simple subset is selected, and the training is carried out until the turning point where the model performance progresses rapidly and levels off, and then samples of increasing difficulty are added to continue the training until the training is carried out until convergence on the complete dataset. In summary, the main contributions of this paper include the following three parts:
\begin{itemize}
\item[1] A teacher model-based difficulty evaluation method is proposed to construct an easy-to-hard training subset using the consistency of the steganographic quality of the samples under multiple teacher models as the difficulty score.
\item[2] A knee point-based training scheduling strategy is proposed to reduce the likelihood of the model falling into overfitting on small training sets and accelerate the training process.
\item[3] Experiments on three large public datasets, ALASKA2, VOC2012, and ImageNet, show that the proposed training strategy is higher than the baseline in several steganographic quality assessment metrics and decoding accuracy metrics, with generalization and validity, while generating steganographic images with low steganalysis scores.
\end{itemize}

\section{Related Work}

\subsection{Deep learning steganography}

The rapid development of deep learning-based steganography analysis models has led to an unprecedented bottleneck in traditional image steganography, and researchers have attempted to introduce deep learning into the field of steganography, and various new types of steganography models have continuously emerged. For example, generative adversarial networks are used to model the complex dependencies between different pixels of an image, so as to generate steganographic images that are more suitable for steganography and more realistic. Volkhonskiy et al \cite{volkhonskiy2016generative} proposed Steganographic Generative Adversarial Networks model (SGAN) in 2016 to generate as realistic as possible carrier images by generative adversarial networks using random noise as input and image steganography by ±1 embedding algorithm. Subsequently, Shi et al \cite{shi2018ssgan} proposed to replace the generative adversarial network with WGAN for generating carrier images that are more consistent with the real distribution based on this foundation. However, this type of steganography algorithm has problems such as unstable training process may lead to the generation of unrealistic images, semantic confusion and other problems. And because the embedding method is still the traditional steganography method, the security has not been greatly improved compared with the traditional steganography method.

Some steganographic models utilize neural networks to automatically learn the minimum embedding distortion cost and employ coding to transform steganography into a problem of finding a better distortion function. For example, generative adversarial networks are utilized to automatically learn the embedding distortion cost and find the embedding location with the minimum distortion cost to reduce the distortion caused by the embedding information to the original image. Tang et al \cite{tang2017automatic} proposed ASDL-GAN, where the generator generates an embedding alteration probability map from the original image, puts the probability map into an embedding simulator (TES) to simulate secret data embedding, generates an alteration location mapping map, and generates a secret-containing image by performing a point-and-point summation of the original image and the alteration location mapping map. Or by iteratively adjusting the embedding domain and error correction ability to prioritize the low-frequency DCT coefficient region, the information embedded in the low-frequency region to generate carrier image simulation compression, the extraction accuracy is not satisfied then modify the error correction ability or adjust the embedding domain to the high-frequency region, so that the processed image can be adapted to the lossy operation of the channel \cite{duan2023robust}. In addition to this, researchers have utilized adversarial attacks and adversarial noise to spoof deep learning based steganalysis models for enhancing the security of steganographic schemes. For example, using multi-granular gradient information and noise residual features to describe texture regions, adaptively adding perturbations to the original image \cite{luo2023improving,luo2023reversible}, and using adversarial perturbations to spoof steganalysis models.

Subsequent researchers have utilized the idea of adversarial training of generative adversarial networks to improve the steganographic image's resistance to steganalysis by using steganography and steganalysis networks as opposites, and training them against each other. The earliest SteGAN based on encoding-decoding network was proposed by Hayes et al \cite{hayes2017generating} for image steganography, which defines a tripartite adversary of Alice, Bob and Eve, representing the image steganography-information extraction-steganalysis process respectively. Alice generates a carrier image and a random n-bit binary secret message as input to generate a carrier image and passes the carrier image to Bob to extract the secret message from it, Eve confirms the presence of the secret message in the image during the training process. Wang et al \cite{wang2018sstegan} added a Dev-square to SteGAN to shrink the distance between the encrypted image and the original carrier image through the adversarial training of Alice and Dev-square, prompting the model to generate more realistic encrypted images. The subsequent SteganoGAN proposed by Zhang et al \cite{zhang2019steganogan} has become the current mainstream adversarial image steganography modeling framework based on encoding-decoding, including the three-party confrontation of encoding-decoding-evaluation parties. Firstly, the embedded information is transformed into binary data with tensor size of and spliced with the image in depth, the encoding network encodes it into the natural image with size of, the information is reconstructed from it by the decoding network, and the evaluating network is used to evaluate the performance of the encoding network in order to generate a more realistic steganographic image.

Adversarial deep learning image steganography algorithms in recent years have focused on steganography robustness optimization research, model structure innovation and loss function optimization and other directions. The research for robust steganography can be categorized into attack simulation enhancement, frequency domain transformation and adversarial samples. For example, by adding content-aware noise projection \cite{xu2022robust} to enhance the robustness of the carrier image for processing containing Gaussian noise, Poisson noise and JPEG compression, and the carrier enhancement module is used to eliminate the impact of the noise of the carrier image and the distortion of the JPEG compression; the Compression Approximation Network (ComNet) is chosen to simulate the JPEG compression operation through self-supervised learning \cite{rao2022towards}; a noise model \cite{bui2023rosteals} is added between the encoding network and decoding network to simulate a variety of common noise attacks; text region segmentation and watermark region localization \cite{chekatamala2022analysis} are used to combat image cropping attacks, and so on. Network (ComNet) is used to simulate the JPEG compression operation through self-supervised learning \cite{rao2022towards}; a noise model is added between the encoding network and the decoding network \cite{bui2023rosteals} to simulate a variety of common noise attacks; and text region segmentation and watermark region localization \cite{chekatamala2022analysis} are utilized to combat the image cropping attack, and so on. This type of method mainly simulates various attacks during the information embedding process, which prompts the embedding network to generate enhanced samples that can resist various common attacks, allowing the decoding network to be trained under various data-enhanced conditions, and realizing that the information can be accurately extracted despite the inclusion of various common noise conditions.

For the study of adversarial image steganography model structure, researchers have introduced reversible neural networks \cite{lu2021large} into steganography to model secret image recovery as a reverse process of image embedding; and introduced flow structure \cite{xu2022robust} to optimize reversible neural network steganography scheme to improve model performance and reduce computational and storage overhead. Optimization of information representation for adversarial steganography is studied, such as proposing layered adversarial training adding subnetworks and discriminators \cite{chen2023layerwise} at each layer of the coding network for capturing the representational capabilities of these layers, adding pre-enhanced and post-enhanced reversible neural network \cite{yang2024pris} structures for improving sample robustness, and giving multiple steganalysis losses to improve security using U-net structures and multiple steganalysis networks \cite{ma2023enhancing}. Subsequently Yang et al \cite{yang2023acgis} proposed to use Siamese Networks to generate adversarial samples to ensure the visual quality of steganographic images by preserving the noise residual relationship in image sub-regions, and adding steganalysis networks for adversarial training to improve security.

\subsection{Curriculum Learning}

The concept of curriculum learning was first introduced by Bengio et al \cite{bengio2009curriculum} in 2009, which advocates that a training strategy that moves from simple to difficult samples can accelerate the convergence of training to a global minimum, and views curriculum learning as a continuation method for global optimization of non-convex functions, arguing that curriculum learning is effective because it can spend less time on noisy and hard to de-train data in the early stages of training, and at the same time, it can guide the training towards better local optima and better generalization effects.

Researchers regard curriculum learning as a continuation method for global optimization of non-convex functions, and believe that curriculum learning is effective because it can spend less time on noisy and training-difficult data at the early stage of training, and at the same time, it can guide the training towards better local optimums and better generalization effects. Curriculum learning is widely used in computer vision \cite{kumar2011learning}, natural language processing \cite{platanios2019competence}, reinforcement learning \cite{saito2018curriculum}, medical diagnosis \cite{zhao2022diagnosing}, etc. By reasonably applying the curriculum learning method for model training can accelerate the model convergence speed, improve the model generalization ability, alleviate the problem of data imbalance, and reduce the negative impact of noisy samples on the model.

The curriculum learning advocates starting with simple samples and progressing gradually to complex samples and knowledge. Three rules are followed in this: the diversity and information (complexity) of the training set is gradually increased, the size of the training set is gradually increased, and eventually the entire data set is used for training. With the development of research, in the process of application researchers have given a broader definition of Curriculum learning so that it can be applied to a wider and wider range of target tasks and domains, such as always using only a fixed-size training set for training \cite{cirik2016visualizing}, starting the process from a highly relevant task \cite{pentina2015curriculum}, training from an unbalanced to a balanced training subset \cite{wang2019dynamic}, and training in the order of simple samples to representative samples \cite{jiang2014self}, etc. For example, in the voiceprint recognition task \cite{heo2022self} gradually increasing the number of samples while increasing the noise during the training process, more speakers are included in the speech, and so on.

In the field of cyberspace security, Ye et al \cite{ye2017deep}, in the steganalysis task, proposed to first train a network on a dataset generated at a higher embedding rate, and then fine-tune the network on another dataset generated at a relatively lower embedding rate; Lee et al \cite{lee2020deep}, in the audio steganalysis task, proposed to train a model from a high BPS (bits per sample) dataset to a low BPS dataset to train the model. In the audio steganography task, Bernat et al \cite{bernat2022using} argued that lower dry-wet (ratio of dry to reverberant signals) parameters represent easier situations, proposing to sample the dry-wet parameter in [0, dry-wet limit], and to implement a training mechanism that gradually increases the difficulty of the Curriculum learning by gradually increasing the dry-wet limit.

\section{STCL}

Curriculum learning mimics the basic idea of gradual progression in the process of human learning curriculum, advocating that the model learns in the order of easier data to more difficult data. Therefore, curriculum learning first needs to assess the difficulty of the dataset, achieve the ordering or division of the data subsets from easy to difficult, and achieve the optimized training for curriculum learning through certain training scheduling rules.The STCL framework consists of two phases: firstly, the difficulty of the carrier image samples is assessed, and then the samples with different difficulty levels are trained in accordance with the corresponding strategies.

\subsection{Difficulty Evaluation Strategies Based on Teacher Models}

\begin{figure}[h]
	\centering
	\centering
	\includegraphics[width=0.9\linewidth]{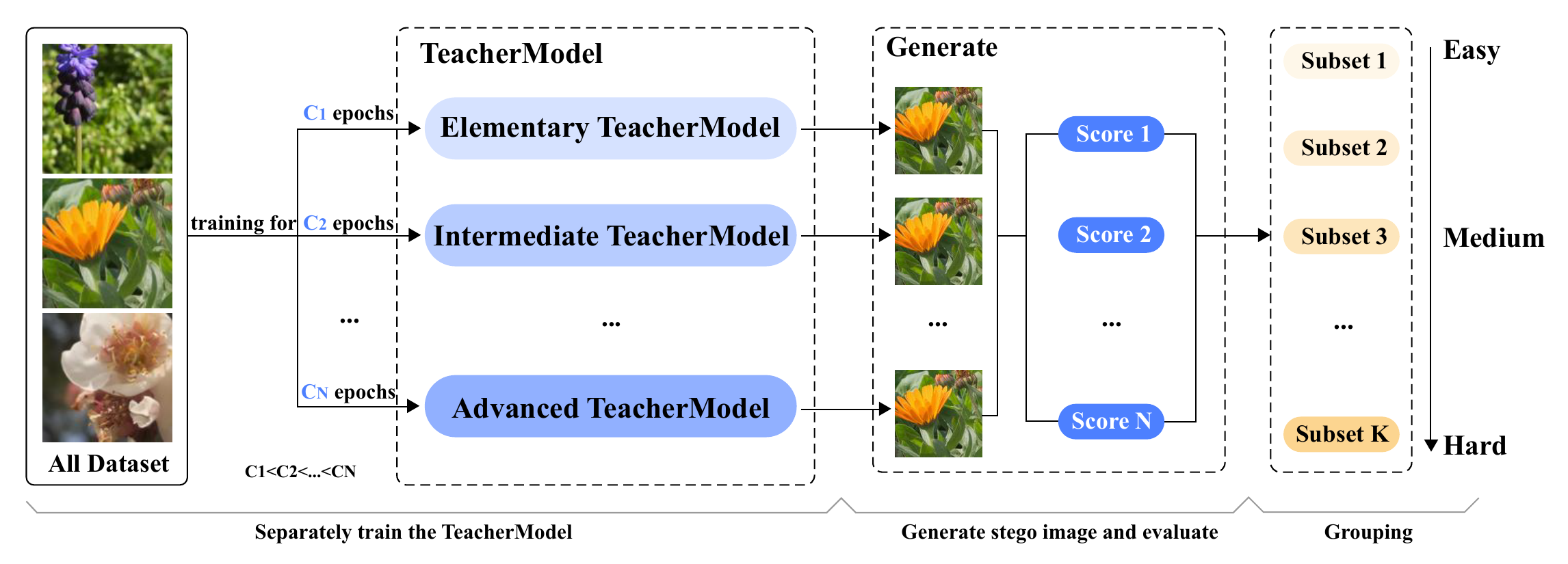}
	\caption{Difficulty Evaluation Strategies Based on Teacher Models.}
	\label{fig1}
\end{figure}

In practice, there are fewer studies on the image difficulty associated with image steganography tasks. Intuitively, information is usually hidden in the complexity of the image texture or at the edges of the image, and it has been proposed that the embedding costs corresponding to different pixel points of a carrier image are different in adaptive steganography-related studies. For example, in an image containing a sky and jagged rocks, modifying the sky pixel points with uniform colors brings more impact to the image than modifying the rock pixel points with complex textures, so for images, embedding information in places with different textures or objects brings different impacts. However, the workload of manually classifying or scoring images for objects or textures is undoubtedly huge, and the image texture complexity metrics currently used for carrier image selection include local variance \cite{bohme2005assessment}, information entropy \cite{gul2011new}, linear prediction error method \cite{huang2016novel}, and wavelet domain model, etc., but such metrics all do the analysis of pixel values or pixel differences, which cannot satisfy the need of designing the neural network applicable to the training difficulty calculations. In this part of the work, a difficulty assessment method based on the teacher model is designed for the image steganography task, as shown in Figure.\ref{fig1}.

The sample evaluation method simulates the steganography state of each image in the actual training through the teacher model, and indirectly outputs the quality score of the steganographic image from the image steganography index as the difficulty score of the sample, which intuitively reacts to the learning difficulty of each image for the model during different periods of the actual training. The teacher model is a three-way adversarial model with the same structure as the actual steganography model. According to the intuition of human brain's learning cognition: for a certain topic, when learning new knowledge, when encountering this question for the first time and can't do it, it may be that this topic is more novel or the knowledge learned is not enough to solve the problem; when doing two sets of exercises and encountering this question again and still can't do it, it indicates that this question is more difficult; when doing a lot of exercises and reviewing, and then encountering this question again and still can't do it, it indicates that the difficulty of this question is too great.  On the other hand, when the same question can be done correctly either for the first time or after review and consolidation, it indicates that the question is simple. For the model, after multiple training sessions with the same batch of samples, the fact that a particular sample still performs poorly on a model of teachers at different levels indicates that the sample is difficult for this model, whereas the fact that a particular sample performs as well as it always has on a model of teachers at different levels indicates that the sample is simple.

The specific method is as follows: individually train multiple teacher models with the same structure as the steganographic network, each image is used as the input of the teacher model, and after the teacher model outputs a secret-containing image containing secret information, its steganographic image quality score is calculated as an indicator of the difficulty of quality steganography. The complete training set $x_{i=1,2,.... ,m}$ is used to train different levels of teacher models, such as teacher model $T_1$ is trained with the complete training set for $C_1$ rounds, teacher model $T_2$ is trained with the complete training set for $C_2$ rounds, and teacher model $T_3$ is trained with the complete training set for $C_3$ rounds, where:

\begin{equation}
C_1 < C_2 < C_3 < C_N
\end{equation}

where $C_N$ represents the number of training sessions in which the model has reached convergence. The teacher model $T_1T_2T_3$ is obtained. Take the sample $x_i \in D$ as the input of the teacher model $T_j$ to get the steganographic image containing the secret information, and evaluate the quality of the steganographic image by using the SSIM and PSNR metrics, as shown in equation (2):

\begin{equation}
S_{ij} = SSIM(T_j(x_i))_{i=1,2,...,m;j=1,2,3}
\end{equation}

\begin{equation}
	P_{ij} = PSNR(T_j(x_i))_{i=1,2,...,m;j=1,2,3}
\end{equation}

$S_{ij}$, $P_{ij}$ denote the scores obtained from the sample $x_i \in D$ after passing through the teacher's model $T_j$ at the assessment indicator $M$, which serves as the basis for difficulty assessment division. Where $M$ is:

\begin{equation}
x_i = 
\begin{cases}
	Easy, & if(S_{ij}(x_i,x_i) \geq \alpha_1 \text{ and } P_{ij}(x_i,x_i) \geq \mu_1 ) \\
	Hard, & if(S_{ij}(x_i,x_i) \leq \alpha_2 \text{ or } P_{ij}(x_i,x_i) \leq \mu_2 ) \\
	Medium, & else
\end{cases}
\end{equation}

$\alpha, \mu $ refers to the structural similarity index (SSIM) and peak signal-to-noise ratio (PSNR) metrics thresholds for subgrouping, which are set based on different problems. Where $\alpha_1 > \alpha_2, \mu_1 > \mu_2$ is satisfied.

\begin{figure}[h]
	\centering
	\centering
	\includegraphics[width=1\linewidth]{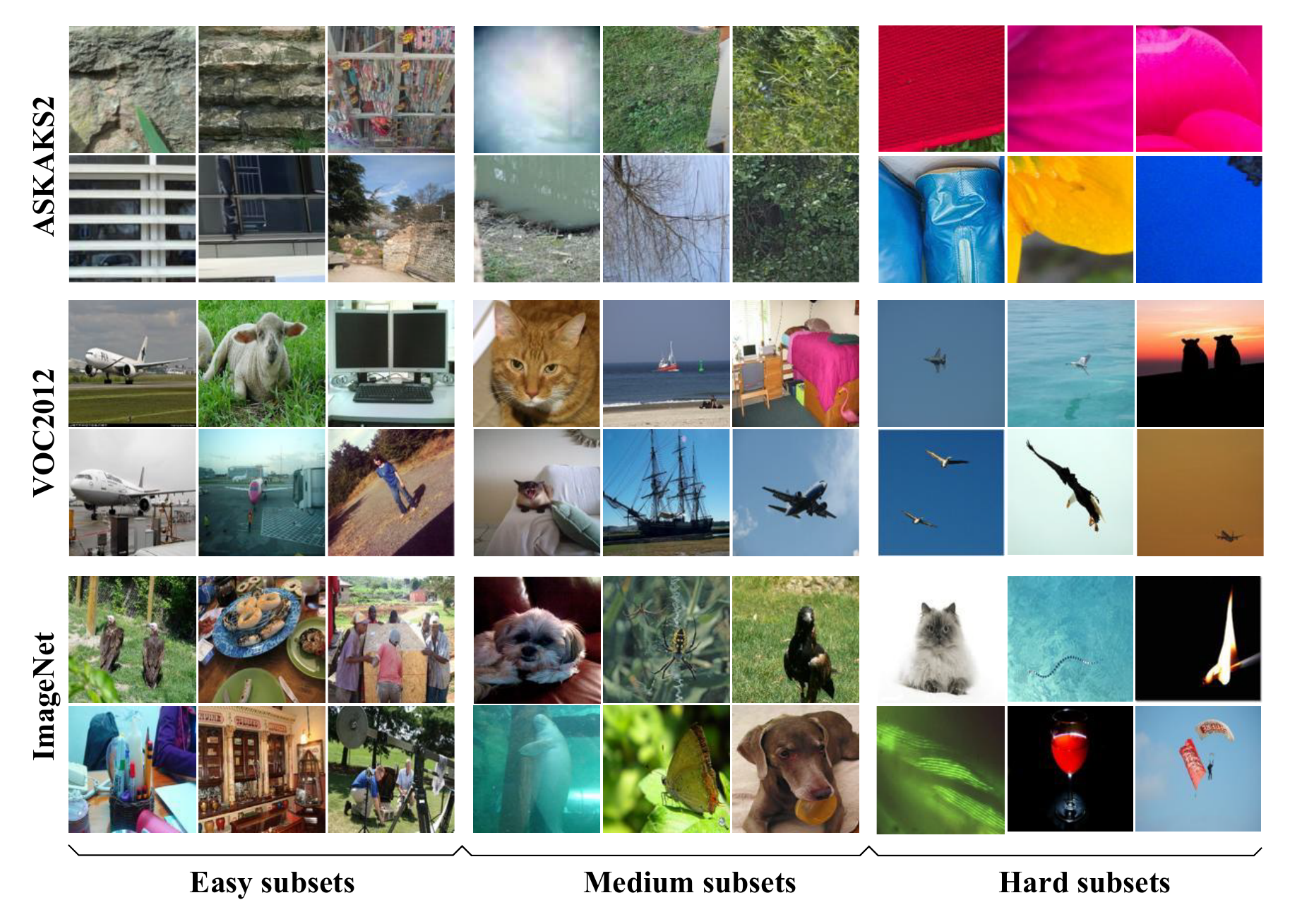}
	\caption{Training subsets with different difficulties obtained by the teacher model difficulty evaluation method.}
	\label{fig2}
\end{figure}

After completing the difficulty evaluation exercise, the complete training set was divided into three training subsets based on the difficulty scores derived from the teacher model, which were categorized as easy, medium, and difficult. The easy subset contained images that performed well and consistently on several different levels of the teacher model, with scores that were all within the same high score range. Conversely, the difficult subset contains images that perform differently on different levels of teacher models and at least one score lies within the low score range. Between easy and difficult is the medium subset. The training subsets obtained from testing on the three datasets are shown in Figure 2. As can be seen in Figure \ref{fig2}, the simple training subset obtained from the teacher model contains more complex textures, while the difficult training subset contains more large color blocks, which is consistent with our intuition that the model is easier for learning images containing more complex textures.

\subsection{Knee point-based training scheduling strategy}

\begin{figure}[h]
	\centering
	\centering
	\includegraphics[width=0.7\linewidth]{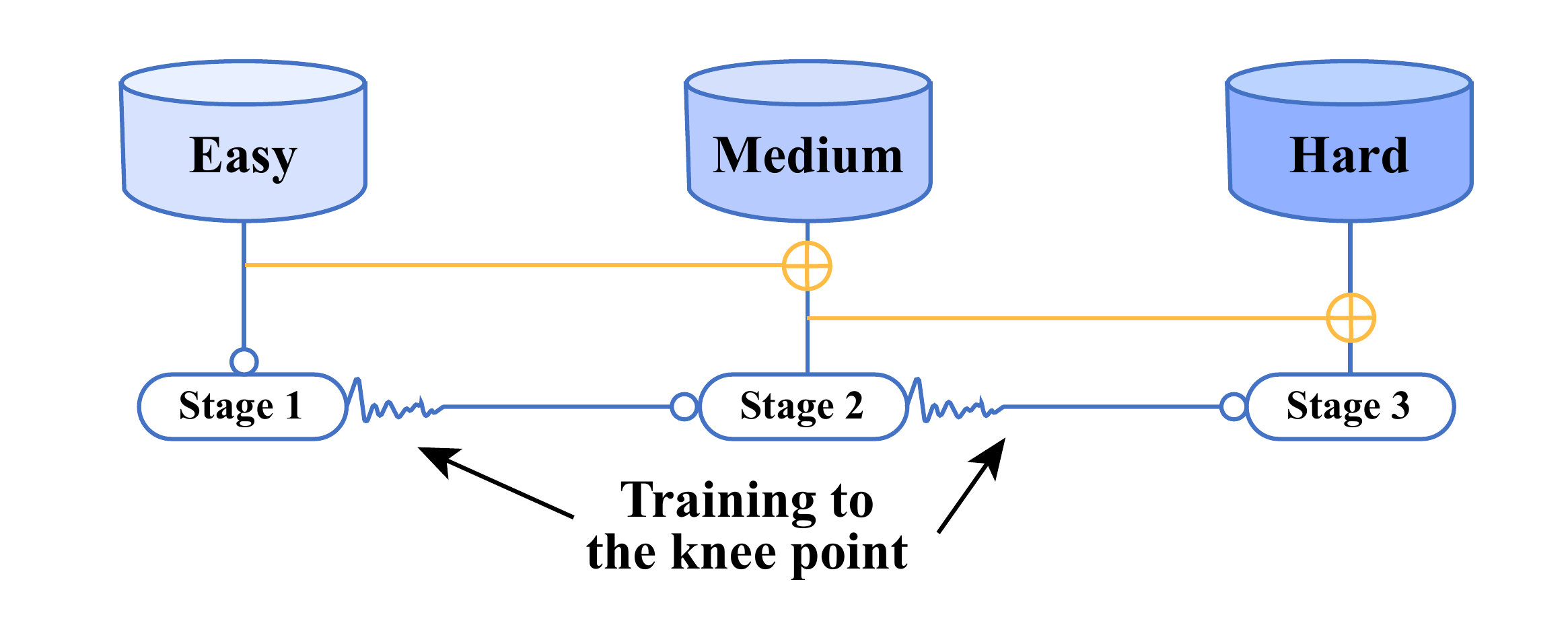}
	\caption{Multi-stage scheduling rules based on knee points.}
	\label{fig3}
\end{figure}

Difficulty evaluator to complete the difficulty score evaluation of the samples need to be further processed, must be specified a reasonable training scheduling rules used to guide the model learning, to achieve the order from easy to difficult gradually added to the model's training set. For the task of steganography on images, a multi-stage scheduling rule based on inflection points is designed. Specifically, the training is divided into three stages as shown in Figure \ref{fig3}.

In the first stage, only a simple subset is used as the training set. This subset contains images that are easier for the initial model to learn, allowing the model to learn the underlying knowledge structure of the data from a large number of simple samples, giving the model a good starting point for initialization and laying the foundation for subsequent learning of more complex and difficult images. In the process of model training, there is a turning point where the model makes rapid progress and tends to converge smoothly, which is called the Knee Point. When the model is trained to an knee point on the simple subset, the training in that stage is stopped. Due to the small number of samples in the easy subset, training on this subset for too long will easily cause the model to fall into a local optimum, and it is easy to see a significant drop in performance when entering the next stage of training. At the same time, stopping training at the inflection point allows the model to roughly grasp the basic knowledge of the data structure, accelerating the training process.

In the second stage, the medium subset is added to the training set and the model is trained to the inflection point on the training set containing the simple and medium subsets. At this point the model learns more discriminative and multi-element features from them to improve image steganography performance. After the first two phases, the model has sufficient underlying knowledge.

Difficult subsets are added in the third stage so that the model is trained to convergence on the full dataset, reviewing the samples that were not adequately learned in the first and second stages. This strategy uses only easy subsets initially during the training process and adds subsets of increasing difficulty one by one, mixing them with the subsets that have already been trained to convergence in the previous stage.

Training to the knee point in the first and second stage can make the model master the general knowledge of this training set, and continue to review and consolidate this part of the samples in the next stage, accelerating the training process. In addition, research \cite{guo2018curriculumnet} shows that difficult, noisy samples are helpful in improving the generalization ability and overall performance of the model. In the steganography task, the effective use of these samples can improve the quality of the model in difficult image steganography that contains large solid color regions and fewer texture edge regions.

\section{Experiments}

\subsection{Experimental Platform and Datasets}

The experiments in this paper are all implemented in Linux system environment using Pytorch 1.10.1 deep learning framework, and the system GPU is NVIDIA GeForce RTX 3090. The datasets use three publicly available large datasets, ALASKA2, Pascal VOC2012, and ImageNet. VOC2012 is a dataset for target detection and semantic segmentation, from which 13k images are selected to form the training set, and the remaining 5k are used as the test and validation sets. ImageNet is a large public computer vision dataset from which 25k images were extracted, of which 20k were used as a training set and the rest were used for testing. ALASKA2 is the public dataset of ALASKA2 Image Steganalysis competition on Kaggle platform. 10k original images are selected as the training set, 3k as the validation set, and 7k as the test set in the “Cover” of ALASKA2 dataset. Due to computational arithmetic limitations, all the original images of the dataset are processed to 128×128 pixels by Matlab program.

\subsection{Parameters}
The experiments in this paper use the Adam provided by Pytorch platform to optimize the network, the initial learning rate (Learning rate) is 0.001, and the momentum parameters (betas) are set to (0.9, 0.999). The number of samples selected for each training (Batchsize) is 8, and the maximum number of iterations of the model (max\_iter) is 120. The base model is an encoding/decoding network based on convolutional modules. The encoding network contains 9 layers of convolutional modules, and the decoding network contains 5 layers of convolutional modules, each of which contains a $3 \times 3$ convolution, a BN, and an activation function LeakyReLU. 

The loss function consists of encoding loss and decoding loss. The encoding loss is evaluated using the SSIM, MSSSIM and RMSE, with the corresponding scale factor of 0.5:0.5:0.3. Binary cross entropy is used for decoding loss. The ratio of encoding loss:decoding loss in the loss function is 1:0.7. The hidden writing capacity in the experiment is D=1-3bpp (i.e., the hidden tensor in a $128 \times 128$ image is $128 \times 128 \times D$ of information). In addition, the parameters $\alpha_1, \alpha_2, \mu_1, \mu_2$ in the difficulty assessment metrics were tested and verified in several datasets, and the experiments in this chapter take $\alpha_1, \alpha_2 = 0.9,0.8, \mu_1, \mu_2 = 20,12$.

\subsection{Experimental results}
1) Training strategy optimization. The sample random training strategy is used as the baseline to compare with the proposed curriculum learning optimization strategy algorithm. Where “Nocl” refers to the baseline scenario using sample random training and “CL” refers to the scenario using STCL training strategy. Table \ref{tab1} shows the test results of the random training strategy and the training based on the proposed curriculum learning optimization strategy on three datasets with 1-3 bpp hidden writing capacity. As can be seen from Table \ref{tab1}, the model performance with the STCL strategy outperforms the baseline scheme with random training in terms of SSIM, MSSSIM, PSNR and RMSE metrics. The secret message reconstruction accuracy is slightly higher than the baseline scheme at 1-2bpp steganographic capacity, indicating that the training strategy can improve the quality of steganography while still maintaining good decoding accuracy.

\begin{table}[htbp]
	\centering
	\caption{Training strategy comparison experiment}
	\label{tab1}
	\renewcommand{\arraystretch}{1.3}
	\begin{tabular}{ccccccccc}
		\hline
		DataSet & D & Scheme & SSIM & MSSSIM & PSNR & RMSE & Accuracy \\
		\hline
		\multirow{6}{*}{ALASKA2} 
		& \multirow{2}{*}{1} & Baseline & 0.9831 & 0.9977 & 33.788 & 0.020 & 0.99 \\
		&  & STCL & \textbf{0.9972} & \textbf{0.9989} & \textbf{37.240} & \textbf{0.013} & 0.99 \\
		\cline{2-8}
		& \multirow{2}{*}{2} & Baseline & 0.9954 & 0.9990 & 35.086 & 0.017 & 0.99 \\
		&  & STCL & \textbf{0.9962} & \textbf{0.9991} & \textbf{36.702} & \textbf{0.014} & 0.92 \\
		\cline{2-8}
		& \multirow{2}{*}{3} & Baseline & 0.9948 & 0.9987 & 34.857 & 0.018 & 0.82 \\
		&  & STCL & \textbf{0.9952} & \textbf{0.9990} & \textbf{38.003} & \textbf{0.012} & 0.81 \\
		\hline
		\multirow{6}{*}{VOC2012} 
		& \multirow{2}{*}{1} & Baseline & 0.9761 & 0.9971 & 32.134 & 0.024 & 0.99 \\
		&  & STCL & \textbf{0.9960} & \textbf{0.9993} & \textbf{38.143} & \textbf{0.012} & 0.99 \\
		\cline{2-8}
		& \multirow{2}{*}{2} & Baseline & 0.9932 & 0.9992 & 35.976 & 0.016 & 0.99 \\
		&  & STCL & \textbf{0.9934} & 0.9992 & \textbf{36.807} & \textbf{0.014} & 0.99 \\
		\cline{2-8}
		& \multirow{2}{*}{3} & Baseline & 0.9940 & 0.9990 & 36.546 & 0.015 & 0.92 \\
		&  & STCL & \textbf{0.9952} & \textbf{0.9994} & \textbf{37.432} & \textbf{0.013} & 0.83 \\
		\hline
		\multirow{6}{*}{ImageNet} 
		& \multirow{2}{*}{1} & Baseline & 0.9901 & 0.9991 & \textbf{36.612} & \textbf{0.014} & 0.96 \\
		&  & STCL & \textbf{0.9953} & 0.9991 & 36.335 & 0.015 & \textbf{0.98} \\
		\cline{2-8}
		& \multirow{2}{*}{2} & Baseline & \textbf{0.9934} & 0.9984 & 33.613 & 0.021 & \textbf{0.99} \\
		&  & STCL & 0.9922 & \textbf{0.9993} & \textbf{38.136} & \textbf{0.012} & 0.73 \\
		\cline{2-8}
		& \multirow{2}{*}{3} & Baseline & 0.9909 & 0.9985 & 33.775 & 0.020 & 0.78 \\
		&  & STCL & \textbf{0.9938} & \textbf{0.9990} & \textbf{36.372} & \textbf{0.015} & \textbf{0.83} \\
		\hline
	\end{tabular}
\end{table}

2) Comparison of training strategies by stages. In order to verify the effectiveness of the proposed Knee point-based training scheduling strategy, the models with different stages of training and randomized training were tested separately, including the three-stage model based on knee point and the baseline with randomized training. Among them, “NoCL” refers to the baseline scheme where samples are selected for random training; “Stage1” refers to the first stage model where only a easy subset is used for training; “Stage2” refers to the model with a mixture of easy and medium difficulty training subsets based on the “Stage1” model; “Stage3” refers to the model that is trained with the complete data set based on the ‘Stage2’ model. Tables \ref{tab2}, \ref{tab3}, and \ref{tab4} show the test results of the four comparison models on the three datasets, respectively.

\begin{table}
	\centering
	\caption{Results of multi-stage comparison experiments on the dataset ALASKA2}
	\label{tab2}
	\renewcommand{\arraystretch}{1.3}
	\begin{tabular}{ccccccc}
		\hline
		D & Scheme & SSIM & MSSSIM & PSNR & RMSE & Accuracy \\
		\hline
		\multirow{4}{*}{1} 
		& Baseline & 0.98351 & 0.99771 & 33.788 & 0.020 & \textbf{0.99} \\
		& Stage1 & 0.99070 & 0.99692 & 31.817 & 0.025 & \textbf{0.99} \\
		& Stage2 & 0.99562 & 0.99809 & 35.932 & 0.016 & \textbf{0.99} \\
		& Stage3 & \textbf{0.99726} & \textbf{0.99894} & \textbf{37.240} & \textbf{0.013} & \textbf{0.99} \\
		\hline
		\multirow{4}{*}{2} 
		& Baseline & 0.99547 & 0.99909 & 35.086 & 0.017 & \textbf{0.99} \\
		& Stage1 & 0.99033 & 0.99756 & 34.650 & 0.018 & 0.97 \\
		& Stage2 & 0.99375 & 0.99861 & \textbf{36.829} & \textbf{0.014} & \textbf{0.99} \\
		& Stage3 & \textbf{0.99626} & \textbf{0.99912} & 36.702 & \textbf{0.014} & 0.92 \\
		\hline
		\multirow{4}{*}{3} 
		& Baseline & 0.99489 & 0.99877 & 34.857 & 0.018 & \textbf{0.82} \\
		& Stage1 & 0.98964 & 0.99750 & 34.627 & 0.019 & 0.76 \\
		& Stage2 & 0.99403 & 0.99851 & 36.361 & 0.015 & 0.80 \\
		& Stage3 & \textbf{0.99520} & \textbf{0.99909} & \textbf{38.003} & \textbf{0.012} & 0.81 \\
		\hline
	\end{tabular}
\end{table}

\begin{table}
	\centering
	\caption{Results of multi-stage comparison experiments on the dataset VOC2012}
	\label{tab3}
	\renewcommand{\arraystretch}{1.3}
	\begin{tabular}{ccccccc}
		\hline
		D & Scheme & SSIM & MSSSIM & PSNR & RMSE & Accuracy \\
		\hline
		\multirow{4}{*}{1} 
		& Baseline & 0.97616 & 0.99716 & 32.134 & 0.024 & \textbf{0.99} \\
		& Stage1 & 0.99217 & 0.99892 & 35.463 & 0.017 & \textbf{0.99} \\
		& Stage2 & 0.98889 & 0.99875 & 35.313 & 0.017 & \textbf{0.99} \\
		& Stage3 & \textbf{0.99603} & \textbf{0.99936} & \textbf{38.143} & \textbf{0.012} & \textbf{0.99} \\
		\hline
		\multirow{4}{*}{2} 
		& Baseline & 0.99326 & \textbf{0.99922} & 35.976 & 0.016 & \textbf{0.99} \\
		& Stage1 & 0.98683 & 0.99868 & 35.139 & 0.017 & 0.97 \\
		& Stage2 & 0.99222 & 0.99896 & 34.539 & 0.019 & \textbf{0.99} \\
		& Stage3 & \textbf{0.99344} & \textbf{0.99922} & \textbf{36.807} & \textbf{0.014} & \textbf{0.99} \\
		\hline
		\multirow{4}{*}{3} 
		& Baseline & 0.99404 & 0.99908 & 36.546 & 0.015 & \textbf{0.92} \\
		& Stage1 & 0.98455 & 0.99821 & 34.485 & 0.019 & 0.84 \\
		& Stage2 & 0.99091 & 0.99909 & 36.429 & 0.015 & 0.85 \\
		& Stage3 & \textbf{0.99529} & \textbf{0.99944} & \textbf{37.432} & \textbf{0.013} & 0.83 \\
		\hline
	\end{tabular}
\end{table}

\begin{table}
	\centering
	\caption{Results of multi-stage comparison experiments on the dataset ImageNet}
	\label{tab4}
	\renewcommand{\arraystretch}{1.3}
	\begin{tabular}{ccccccc}
		\hline
		D & Scheme & SSIM & MSSSIM & PSNR & RMSE & Accuracy \\
		\hline
		\multirow{4}{*}{1} 
		& Baseline & 0.99017 & \textbf{0.99914} & \textbf{36.612} & \textbf{0.014} & 0.96 \\
		& Stage1 & 0.98112 & 0.99752 & 32.652 & 0.024 & \textbf{0.99} \\
		& Stage2 & 0.99160 & 0.99908 & 35.506 & 0.016 & 0.87 \\
		& Stage3 & \textbf{0.99535} & \textbf{0.99914} & 36.335 & 0.015 & 0.98 \\
		\hline
		\multirow{4}{*}{2} 
		& Baseline & \textbf{0.99348} & 0.99843 & 33.613 & 0.02 & \textbf{0.99} \\
		& Stage1 & 0.98775 & 0.99824 & 33.891 & 0.020 & 0.59 \\
		& Stage2 & 0.98934 & 0.99912 & 36.095 & 0.015 & 0.70 \\
		& Stage3 & 0.99222 & \textbf{0.99937} & \textbf{38.136} & \textbf{0.012} & 0.73 \\
		\hline
		\multirow{4}{*}{3} 
		& Baseline & 0.99091 & 0.99852 & 33.775 & 0.020 & 0.78 \\
		& Stage1 & 0.99116 & 0.99899 & 36.040 & 0.016 & 0.79 \\
		& Stage2 & 0.99280 & \textbf{0.99913} & 36.075 & \textbf{0.015} & \textbf{0.84} \\
		& Stage3 & \textbf{0.99384} & 0.99906 & \textbf{36.372} & \textbf{0.015} & 0.83 \\
		\hline
	\end{tabular}
\end{table}

As can be seen from Tables \ref{tab2}, \ref{tab3}, and \ref{tab4}, the performance of the third stage model using the knee point-based training scheduling strategy is consistently better than that of the baseline scheme using random training. In particular, each stage of the model of the knee point-based training strategy scheme outperforms the previous stage, proving that the proposed curriculum learning is effective because the choice of a simple subset at the beginning of the training period can give the model a good starting point for initialization, guiding the model towards better parameter regions, and reducing the likelihood of difficulty in fitting the model at the beginning of the training period.

3) Difficulty-based training strategy validation. In order to further validate the effectiveness of the curriculum learning training strategy on the image steganography model, multiple teacher models are still chosen to train the dataset to divide the dataset into three training subsets with different levels of difficulty, easy, medium and difficult. Only the easy subset, only the medium subset, and only the subset containing both the easy and medium subsets are chosen as comparison models to verify the effectiveness of the proposed three-stage training strategy from easy to hard, and the experimental results are shown in Table \ref{tab5}. As can be seen from Table \ref{tab5}, the performance of the models trained with only the easy subset and only the medium subset is close to the performance of the model trained randomly, while the performance of the model trained with only the difficult subset is drastically reduced. This shows that in addition to the difference in the number of samples in the dataset, learning from difficult images alone does not enable the model to achieve good performance. The performance of the model selected for random training with a mixture of easy and medium subsets is lower than the performance of the model trained with Curriculum learning, proving the effectiveness of the proposed curriculum learning training strategy.

\begin{table}[htbp]
	\centering
	\caption{Training test results for difficulty subsets}
	\label{tab5}
	\renewcommand{\arraystretch}{1.3}
	\begin{tabular}{ccccc}
		\hline
		Scheme & SSIM & MSSSIM & PSNR & Accuracy \\
		\hline
		Baseline & 0.98351 & 0.99771 & 33.788 & 0.99 \\
		Only-Easy & 0.96453 & 0.99511 & 32.099 & 0.99 \\
		Only-Medium & 0.97995 & 0.99527 & 33.712 & 0.99 \\
		Only-Hard & 0.87476 & 0.94933 & 19.066 & 0.86 \\
		Only-Easy+Medium & 0.98587 & 0.99823 & 36.067 & 0.99 \\
		STCL & \textbf{0.99726} & \textbf{0.99894} & \textbf{37.240} & 0.99 \\
		\hline
	\end{tabular}
\end{table}

\begin{figure}[htbp]
	\centering
	\centering
	\includegraphics[width=0.9\linewidth]{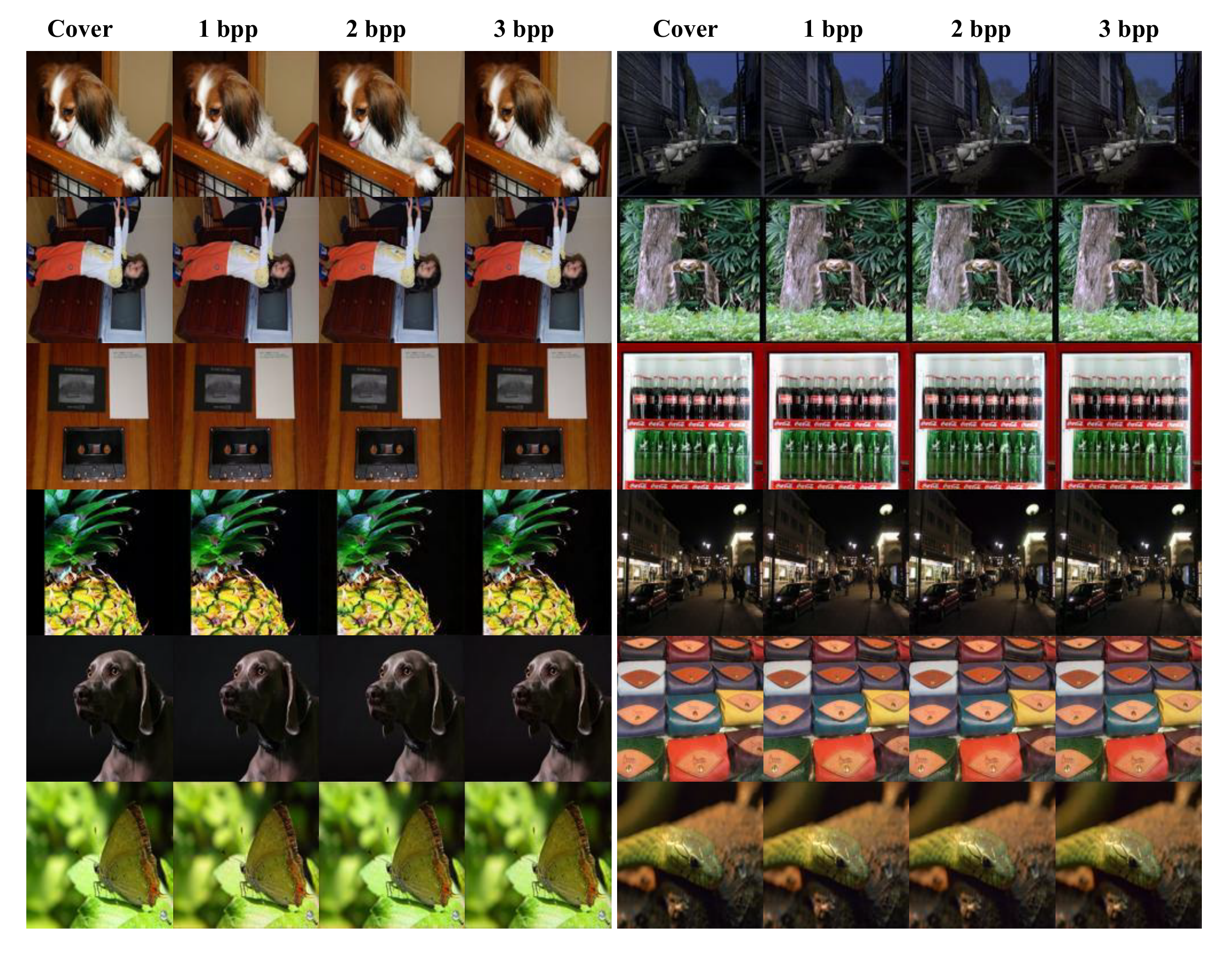}
	\caption{Comparison of cover and stego images under 1-3 bpp capacity steganography.}
	\label{fig4}
\end{figure}

4)  Image Steganography Quality Visual Testing. In order to further validate the steganographic quality of the model, some images from ImageNet, ALASKA2, and VOC2012 test sets were selected for testing. Firstly, the original images under 1-3 bpp steganographic capacity were verified and compared with the steganographic images, as shown in Fig. \ref{fig4}. As can be seen from Figure. \ref{fig4}, the steganographic images at 1-3 bpp steganographic capacity are more similar to the original images in terms of color and brightness, and there is no obvious difference under human visual observation. 

\begin{figure}[htbp]
	\centering
	\centering
	\includegraphics[width=0.7\linewidth]{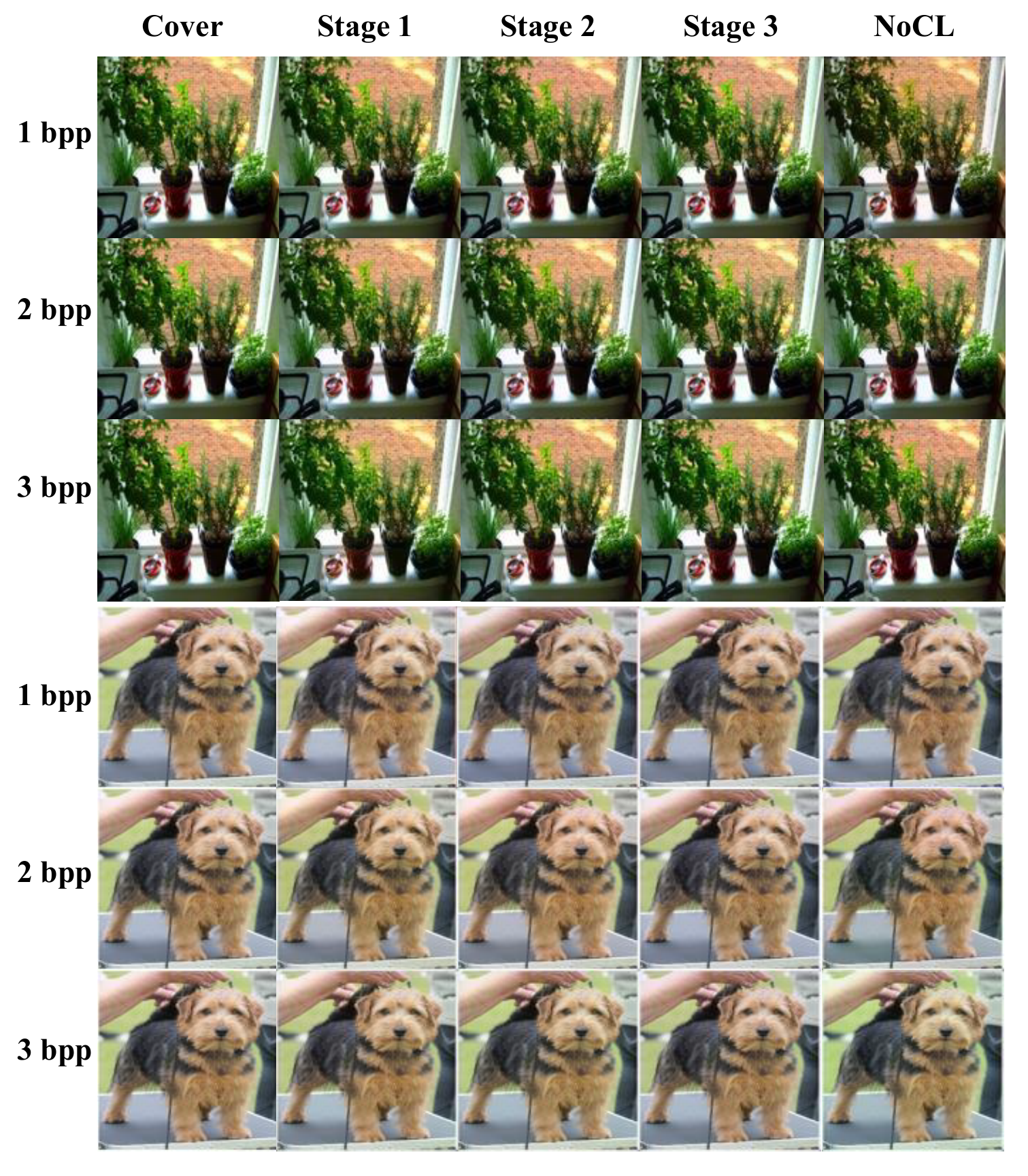}
	\caption{Comparison of stego image generated by multi-stage model and cover image.}
	\label{fig5}
\end{figure}

Subsequently, the knee point-based training strategy was validated, and the baseline models with 1-3 stages and random training were chosen to generate steganographic images and compared with the original images, as shown in Figure. \ref{fig5}. As known from Fig. 5, at 1-3 bpp steganographic capacity, the steganographic image generated by the 1-3 stage model is extremely similar to the original image in terms of color, structure and brightness, while the steganographic image generated using the randomly trained baseline model is slightly yellowish in color.

\begin{figure}[htbp]
	\centering
	\centering
	\includegraphics[width=0.9\linewidth]{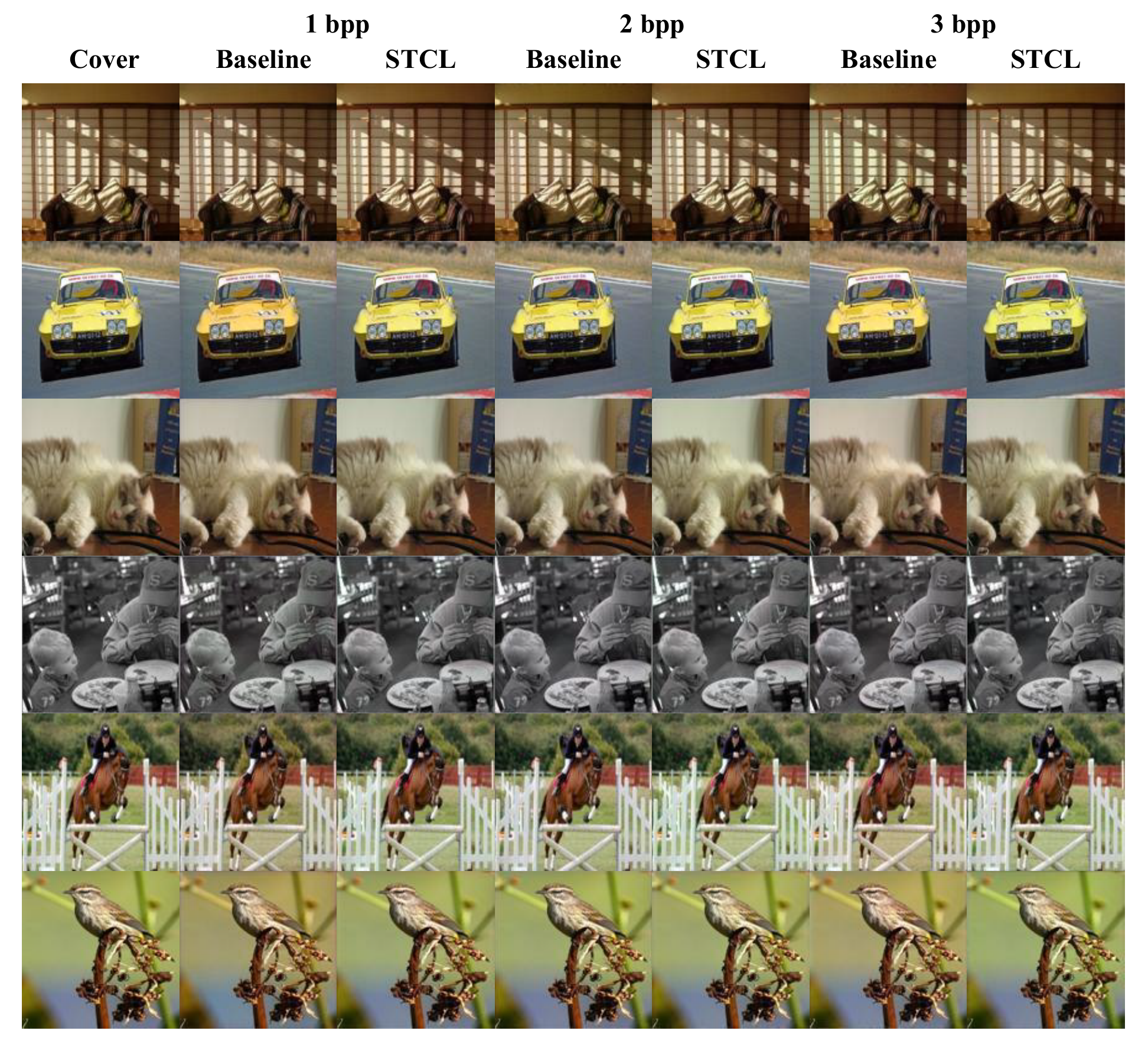}
	\caption{Comparative validation using STCL.}
	\label{fig6}
\end{figure}

In addition to this, multiple steganographic images under 1-3 bpp using the random training model are chosen to be compared with the steganographic images trained using the STCL, as shown in Fig. 6. As can be seen from Figure. \ref{fig6}, the steganographic images generated by the model without the training of the curriculum learning strategy show yellowish and pinkish phenomena on several test images, and there is a more obvious difference between the original image and the original image in color, while the steganographic images generated by the model with the training of the curriculum learning strategy are more similar to the original image in terms of color and brightness.

\begin{figure}[htbp]
	\centering
	\centering
	\includegraphics[width=0.9\linewidth]{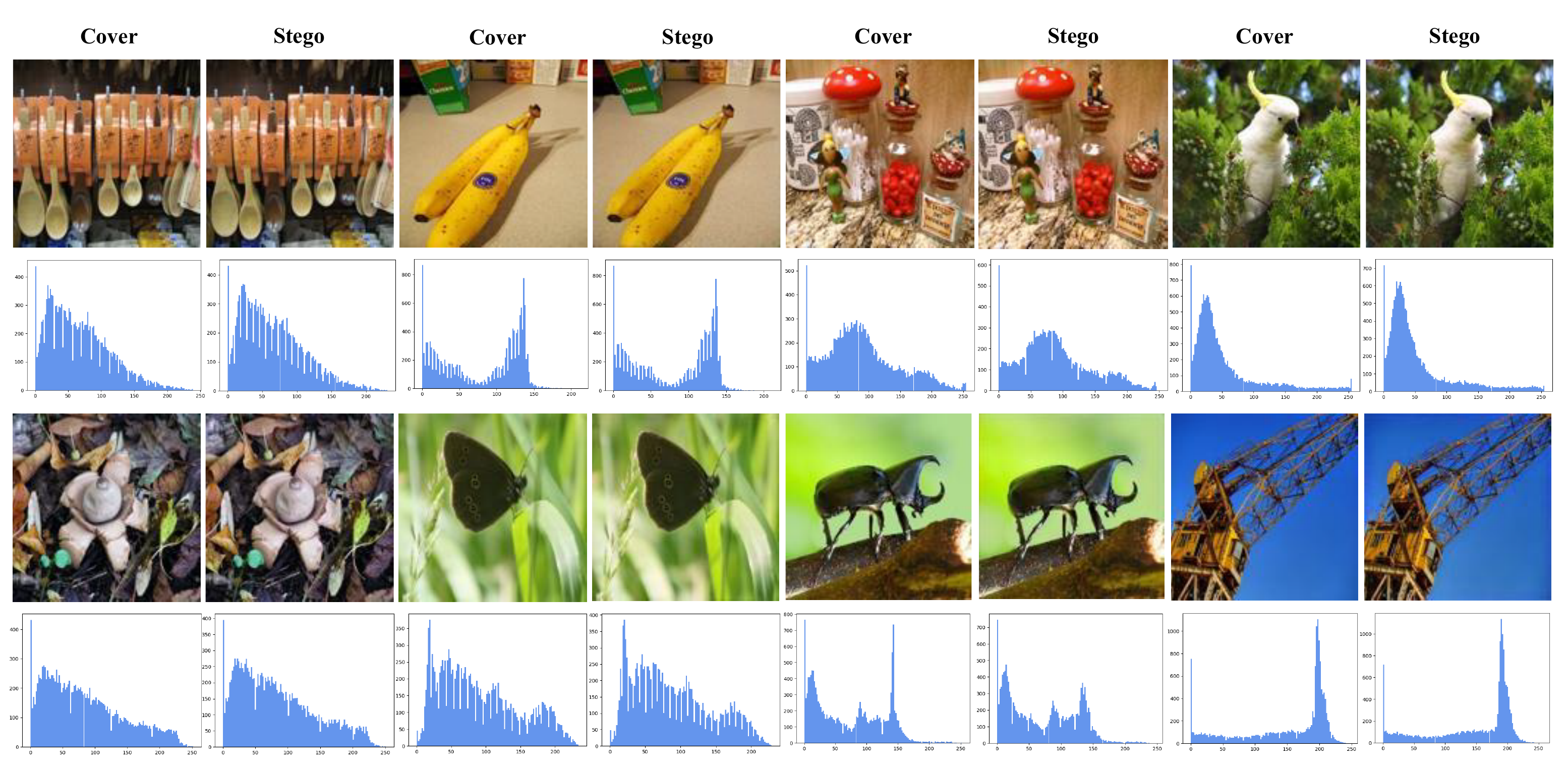}
	\caption{Histogram comparison of the cover and the stego image.}
	\label{fig7}
\end{figure}

To more fully demonstrate the effectiveness of STCL, multiple histograms of the steganographic image versus the original image are plotted, as shown in Figure. \ref{fig7}. It can be observed that the difference between the histograms of the original image and the steganographic image is very small, indicating that the optimization scheme does not destroy the visual integrity of the image.

5)  Comparison of different difficulty images. In order to verify the steganographic performance of the models optimized by the curriculum learning training strategy on different difficulty images, test sets were selected from each of the three datasets for difficulty assessment, which were classified into easy, medium, and difficult subsets. The models with different stages of training and randomized training were tested separately, including the three-stage model based on knee point and the baseline with randomized training, and the test results of 1-3bpp on the three datasets are shown in Table \ref{tab6}.

\begin{table}
	\centering
	\caption{Contrastive experiment on images of varying difficulty levels}
	\label{tab6}
	\renewcommand{\arraystretch}{1.3}
	\begin{tabular}{cccccccc}
		\hline
		Dataset & Subset & Scheme & SSIM & MSSSIM & PSNR & RMSE & Accuracy \\
		\hline
		\multirow{12}{*}{ALASKA2} 
		& \multirow{4}{*}{Easy} 
		& NoCL & 0.98645 & 0.99887 & 33.969 & 0.020 & \textbf{0.99} \\
		& & Stage1 & 0.98645 & 0.99788 & 31.994 & 0.025 & \textbf{0.99} \\
		& & Stage2 & 0.99791 & \textbf{0.99939} & 36.537 & 0.014 & \textbf{0.99} \\
		& & Stage3 & \textbf{0.99846} & 0.99911 & \textbf{37.511} & \textbf{0.013} & \textbf{0.99} \\
		\cline{2-8}
		& \multirow{4}{*}{Medium} 
		& NoCL & 0.98300 & 0.99817 & 33.853 & 0.020 & \textbf{0.99} \\
		& & Stage1 & 0.98982 & 0.99644 & 31.585 & 0.026 & \textbf{0.99} \\
		& & Stage2 & 0.99603 & 0.99893 & 35.545 & 0.016 & \textbf{0.99} \\
		& & Stage3 & \textbf{0.99743} & \textbf{0.99904} & \textbf{36.909} & \textbf{0.014} & \textbf{0.99} \\
		\cline{2-8}
		& \multirow{4}{*}{Hard} 
		& NoCL & 0.95963 & 0.99158 & 33.221 & 0.022 & \textbf{0.99} \\
		& & Stage1 & 0.96537 & 0.99110 & 32.384 & 0.024 & \textbf{0.99} \\
		& & Stage2 & 0.97025 & 0.99144 & 33.723 & 0.021 & \textbf{0.99} \\
		& & Stage3 & \textbf{0.98433} & \textbf{0.99623} & \textbf{37.771} & \textbf{0.013} & 0.98 \\
		\hline
		\multirow{12}{*}{VOC2012} 
		& \multirow{4}{*}{Easy} 
		& NoCL & 0.98102 & 0.99756 & 32.108 & 0.024 & \textbf{0.99} \\
		& & Stage1 & 0.99540 & \textbf{0.99909} & 35.587 & 0.016 & \textbf{0.99} \\
		& & Stage2 & 0.99228 & 0.99897 & 35.368 & 0.017 & \textbf{0.99} \\
		& & Stage3 & \textbf{0.99765} & 0.99943 & \textbf{38.263} & \textbf{0.012} & \textbf{0.99} \\
		\cline{2-8}
		& \multirow{4}{*}{Medium} 
		& NoCL & 0.97202 & 0.99658 & 31.903 & 0.025 & \textbf{0.99} \\
		& & Stage1 & 0.99161 & 0.99877 & 35.170 & 0.017 & \textbf{0.99} \\
		& & Stage2 & 0.98745 & 0.99856 & 35.140 & 0.017 & \textbf{0.99} \\
		& & Stage3 & \textbf{0.99561} & \textbf{0.99927} & \textbf{37.845} & \textbf{0.012} & \textbf{0.99} \\
		\cline{2-8}
		& \multirow{4}{*}{Hard} 
		& NoCL & 0.93777 & 0.99542 & 34.337 & 0.019 & \textbf{0.99} \\
		& & Stage1 & 0.95892 & 0.99782 & 35.919 & 0.016 & \textbf{0.99} \\
		& & Stage2 & 0.95673 & 0.99719 & 35.862 & 0.016 & \textbf{0.99} \\
		& & Stage3 & \textbf{0.97951} & \textbf{0.99900} & \textbf{38.628} & \textbf{0.011} & \textbf{0.99} \\
		\hline
		\multirow{12}{*}{ImageNet} 
		& \multirow{4}{*}{Easy} 
		& NoCL & 0.99552 & 0.99898 & 34.126 & 0.020 & 0.86 \\
		& & Stage1 & 0.98969 & 0.99836 & 33.056 & 0.022 & \textbf{0.99} \\
		& & Stage2 & 0.99657 & \textbf{0.99929} & \textbf{35.880} & 0.016 & 0.87 \\
		& & Stage3 & \textbf{0.99785} & 0.99938 & 36.550 & \textbf{0.015} & 0.98 \\
		\cline{2-8}
		& \multirow{4}{*}{Medium} 
		& NoCL & 0.99064 & 0.99865 & 33.352 & 0.022 & 0.87 \\
		& & Stage1 & 0.98081 & 0.99720 & 32.355 & 0.025 & \textbf{0.99} \\
		& & Stage2 & 0.99249 & \textbf{0.99902} & 35.148 & 0.017 & 0.87 \\
		& & Stage3 & \textbf{0.99561} & 0.99897 & \textbf{35.817} & \textbf{0.016} & 0.98 \\
		\cline{2-8}
		& \multirow{4}{*}{Hard} 
		& NoCL & 0.93172 & 0.99705 & 35.389 & 0.017 & 0.93 \\
		& & Stage1 & 0.90366 & 0.99523 & 30.914 & 0.029 & \textbf{0.97} \\
		& & Stage2 & 0.94439 & 0.99782 & 35.425 & 0.017 & 0.86 \\
		& & Stage3 & \textbf{0.97327} & \textbf{0.99904} & \textbf{37.001} & \textbf{0.014} & \textbf{0.97} \\
		\hline
	\end{tabular}
\end{table}

As can be seen from Table \ref{tab6}, the proposed STCL strategy improves the performance on the simple, medium and difficult subsets on all three datasets. The results on the difficult subset of the three datasets can be observed that the model without the STCL strategy performs poorly on the difficult subset, and each of the hidden writing metrics is lower than the test results on the simple and medium subsets. On the other hand, the model that chooses the STCL strategy still maintains a better steganography performance on the difficult subset, indicating that the reasonable use of images with different difficulties for training can effectively improve the generalization performance of the model.

6) Comparison of training convergence nodes. The knee point based multi-stage training scheduling strategy in the first and second stage selects to stop training at the knee point where the model performance is rapidly progressing with a tendency to converge smoothly. In order to verify the effectiveness of the method, the knee point-based training three-stage model, the random training model, and the model that replaces the first and second stages of training to the knee point with training to convergence are selected for comparison experiments on the ALASKA2 dataset, and the results of the experiments are shown in Table \ref{tab7}.

\begin{table}[htbp]
	\centering
	\caption{Experiments on the effectiveness of Knee point}
	\label{tab7}
	\renewcommand{\arraystretch}{1.3}
	\begin{tabular}{ccccccccc}
		\hline
		& \multicolumn{4}{c}{Knee point} & \multicolumn{4}{c}{Convergence} \\
		\cmidrule(lr){2-5} \cmidrule(lr){6-9}
		Metric & SSIM & MSSSIM & PSNR & Accuracy & SSIM & MSSSIM & PSNR & Accuracy \\
		\hline
		Stage 1 & 0.989 & \textbf{0.997} & 34.62 & 0.99 & \textbf{0.992} & \textbf{0.998} & 34.91 & 0.99 \\
		Stage 2 & 0.994 & \textbf{0.998} & 36.36 & 0.99 & \textbf{0.995} & \textbf{0.998} & 35.03 & 0.99 \\
		Stage 3 & \textbf{0.995} & \textbf{0.999} & \textbf{38.00} & 0.99 & \textbf{0.996} & \textbf{0.999} & \textbf{37.11} & 0.99 \\
		\hline
	\end{tabular}
\end{table}

As can be observed from Table \ref{tab7}, the performance of Stage1 using full convergence is slightly better than the performance of the corresponding model trained to knee point, and the performance of the first stage model trained to knee point is close to the performance of full convergence. The model trained in the second stage based on the fully converged Stage1 model shows a performance degradation in the early stage of training and the model performance at the final convergence is slightly lower than that using the model trained up to the knee point, and the performance of the subsequent Stage3 model is still slightly lower than the performance of the corresponding model trained up to the knee point. The experiments show that the model stops training at knee point between its performance progressing rapidly and leveling off, ensuring that the model learns the basics on that training subset. It also reduces the likelihood of the model falling into overfitting and local optimality due to a small subset of easy data, gives the model a good starting point for initialization, and greatly reduces training time. 

\begin{figure}[htbp]
	\centering
	\centering
	\includegraphics[width=0.8\linewidth]{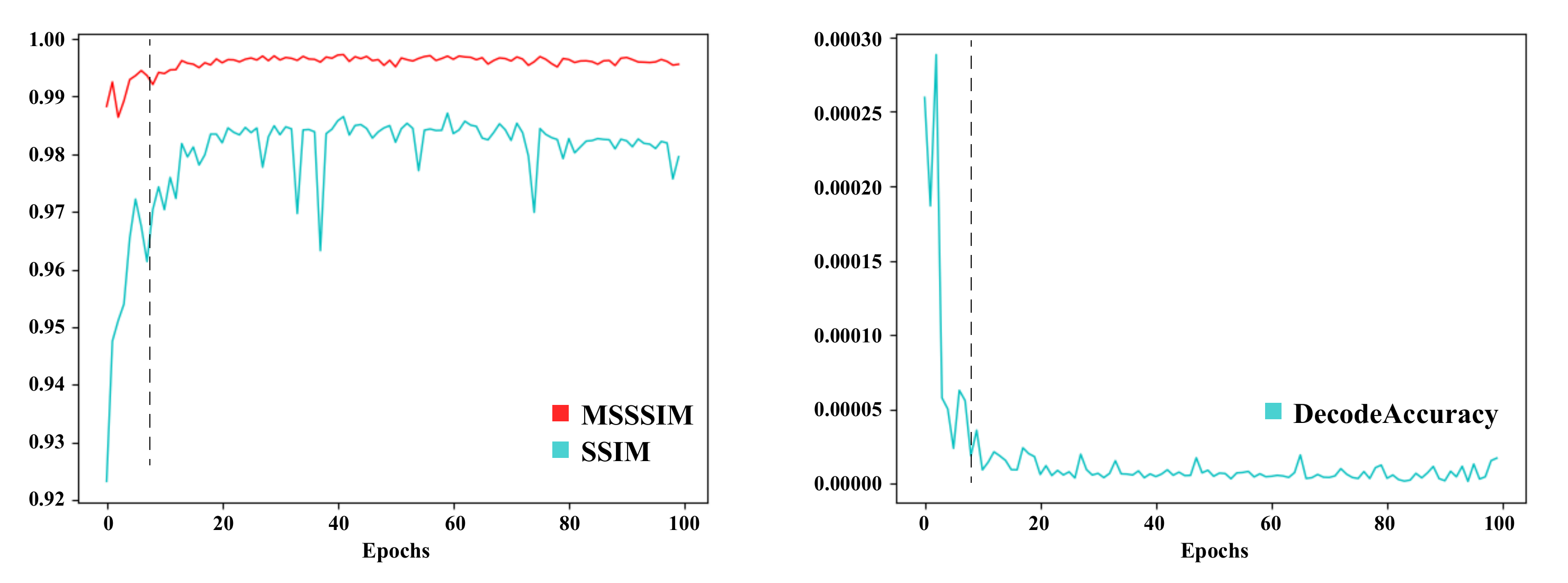}
	\caption{A knee point located between rapid progress and convergence of models.}
	\label{fig8}
\end{figure}

The first stage of the model on the three datasets takes only 15-30 epochs to reach the knee point between fast progress and convergence of the model, while more training time is needed to reach full convergence. Figure \ref{fig8} shows the first stage loss curves for training on the dataset VOC2012, with the dashed line denoting the knee point, which is located at the turning point between the model's rapid progress and convergence. Also, since the model does not undergo complete learning in the first two phases, training to convergence on the complete dataset in the third phase can be compensated for by revisiting the simple and medium subsets to improve model performance.

6)  Security Testing. In order to investigate whether the proposed Curriculum-learning optimization method is effective in improving the security of steganographic images, a steganalysis model is selected to validate the method. The trained XuNet \cite{xu2016structural} model is selected for steganalysis testing, and the randomly trained baseline model, the model with stages 1-3 of the Curriculum learning training strategy is selected to generate steganographic images, which are input to the trained XuNet model to output steganalysis scores. This steganalysis output ranges from [0,1], the closer the value is to 1, i.e., it indicates that there is a high probability that the image contains secret information and low security. The final result uses the average of the test set scores as the final steganalysis score and the results are shown in Table \ref{tab8}. As can be seen from Table 8, the steganalysis scores of the model trained in the STCL strategy are slightly lower than the baseline model trained randomly and the security of the image is improved at each stage as the training progresses.

\begin{table}[htbp]
	\centering
	\caption{The steganalysis results}
	\label{tab8}
	\renewcommand{\arraystretch}{1.3}
	\begin{tabular}{ccccccc}
		\hline
		Dataset & D & NoCL & Stage1 & Stage2 & Stage3 \\
		\hline
		\multirow{3}{*}{ALASKA2}
		& 1 & 0.44 & 0.43 & 0.40 & \textbf{0.36} \\
		& 2 & \textbf{0.39} & 0.40 & 0.45 & 0.40 \\
		& 3 & 0.44 & 0.44 & \textbf{0.38} & 0.39 \\
		\hline
		\multirow{3}{*}{VOC2012}
		& 1 & \textbf{0.43} & \textbf{0.43} & 0.44 & 0.44 \\
		& 2 & 0.45 & 0.44 & 0.42 & \textbf{0.40} \\
		& 3 & \textbf{0.36} & 0.44 & 0.42 & 0.39 \\
		\hline
		\multirow{3}{*}{ImageNet}
		& 1 & 0.36 & 0.35 & \textbf{0.32} & 0.34 \\
		& 2 & \textbf{0.32} & 0.37 & 0.36 & 0.33 \\
		& 3 & 0.41 & 0.42 & 0.40 & \textbf{0.38} \\
		\hline
	\end{tabular}
\end{table}

7)  Method generalization test. In order to further validate the generalization of the proposed training strategy, several models with different structures were selected for testing, keeping the training strategy parameters, datasets and experimental conditions set the same. SteganoGAN \cite{bui2023rosteals} and FC-DenseNet \cite{heo2022self} models are selected for the experiments. In this part of the experiment, the input RGB image in Duan et al.'s study \cite{heo2022self} is modified to binary information, and the rest of the loss function and parameter settings are the same as those of the model in this paper, and comparative experiments are conducted on the ALASKA2 dataset. The results are shown in Table \ref{tab9}.

\begin{table}[htbp]
	\centering
	\caption{Results of experiments on the generality of STCL}
	\label{tab9}
	\renewcommand{\arraystretch}{1.3}
	\begin{tabular}{cccccc}
		\hline
		Model & Scheme & SSIM & MSSSIM & PSNR & Accuracy \\
		\hline
		\multirow{4}{*}{StaganoGAN} 
		& NoCL & 0.98970 & 0.99907 & 37.669 & 0.91 \\
		& Stage1 & 0.99163 & 0.99934 & \textbf{41.306} & 0.83 \\
		& Stage2 & \textbf{0.99701} & \textbf{0.99969} & 40.351 & 0.72 \\
		& Stage3 & 0.98324 & 0.99934 & 39.669 & \textbf{0.95} \\
		\hline
		\multirow{4}{*}{FC-DenseNet} 
		& NoCL & 0.93603 & 0.98954 & 34.268 & 0.61 \\
		& Stage1 & 0.94232 & 0.99725 & 36.828 & - \\
		& Stage2 & 0.96964 & \textbf{0.99876} & 39.561 & - \\
		& Stage3 & \textbf{0.98050} & 0.99794 & \textbf{40.985} & 0.72 \\
		\hline
	\end{tabular}
\end{table}

\section{Conclusion}
In this paper, we propose a curriculum learning training strategy STCL for deep learning image steganography models, including a difficulty assessment strategy based on the teacher's model and an knee point-based training scheduling strategy. The model is trained on easy images only when the fitting ability is poor at the initial stage, and gradually expanded to more difficult images, and the training stop nodes are controlled at each stage of training to accelerate the network convergence process and reduce the possibility of overfitting. The STCL proposed in this paper verifies its excellent performance in improving the quality of steganographic images and enhancing the security on several datasets, not only on the overall test set, but also on the difficult images containing large solid color regions can still show better performance. In addition, through various experimental comparisons, it is proved that STCL has good generality and generalization performance. However, the STCL designed in this paper is only applicable to binary information embedding, and future attempts will be made for image steganography models where the embedding carrier is an image.


\bibliographystyle{unsrt}  
\bibliography{ref}  


\end{document}